\documentclass[letterpaper, 10 pt, conference]{ieeeconf}  % Comment this line out if you need a4paper

\IEEEoverridecommandlockouts                              % This command is only needed if 
\usepackage{color}
\usepackage{amsmath, amssymb, amscd, amsfonts} % assumes amsmath package installed
\usepackage{graphics} % for pdf, bitmapped graphics files
\usepackage{graphicx}
\usepackage{hyperref}
\usepackage{tabularx}
\usepackage{multirow}
\usepackage{booktabs}
\usepackage{cite}
\usepackage{float}
\usepackage{subfig}
\usepackage{caption}
\usepackage{algorithm}
\usepackage{algorithmic}% line52,53包冲突
\usepackage{bbding}
\usepackage{array}
\usepackage{makecell}
\usepackage{color}
\usepackage{amsmath, amssymb, amscd, amsfonts} % assumes amsmath package installed
\usepackage{graphics} % for pdf, bitmapped graphics files
\usepackage{graphicx}
\usepackage{hyperref}
\usepackage{tabularx}
\usepackage{multirow}
\usepackage{vcell}
\usepackage{colortbl}
\usepackage{booktabs}
\usepackage{cite}
\usepackage{float}
\usepackage{subfig}
\usepackage{caption}
\usepackage{algorithm}
\usepackage{algorithmic}
\usepackage{bbding}
\usepackage{array}
\usepackage{makecell}
\usepackage{lipsum}
\usepackage{tikz}
\usepackage{url}
\usetikzlibrary{shapes.geometric, arrows, positioning}

\UseRawInputEncoding

\graphicspath{ {./figs/} }

\title{\LARGE \bf
    ADM: Accelerated Diffusion Model via Estimated Priors for Robust Motion Prediction under Uncertainties
    
}

\author{
    Jiahui Li$^{\dag}$, Tianle Shen$^{\dag}$, Zekai Gu, Jiawei Sun,  Chengran Yuan, \\ Yuhang Han, Shuo Sun, and Marcelo H. Ang Jr.%
    \thanks{
        $^{\dag}$ Indicates Equal Contribution.
    }
    \thanks{
        All authors are with the Department of Mechanical Engineering, National University of Singapore, Singapore 119077 (e-mail: \{zekai.gu, sunjiawei, chengran.yuan, yuhang\_han, shuo.sun,\}@u.nus.edu; mpeangh@nus.edu.sg).
    }%
    \thanks{
        Project page: \href{https://github.com/skygoo2000/ADM}{https://github.com/skygoo2000/ADM}.
    }%
}

\begin{document}

\maketitle
\thispagestyle{empty}
\pagestyle{empty}

%%%%%%%%%%%%%%%%%%%%%%%%%%%%%%%%%%%%%%%%%%%%%%%%%%%%%%%%%%%%%%%%%%%%%%%%%%%%%%%%
\begin{abstract}

% \textcolor{red}{
% Outline
% \begin{itemize}
%     \item Diffusion has demonstrated capabilities in many fields/tasks...
%     \item Existing diffusion-based Motion prediction methods require long denoising steps.
%     \item We propose a two-stage diffusion-based method that predicts agents' future trajectories for autonomous driving. 
%     \item We explicitly learn an explicit prior distribution by leveraging the HD map information and the agent track histories to facilitate the diffusion process, achieving higher sampling efficiency while not compromising prediction accuracy.
% \end{itemize}
% }

% motion prediction is a vital problem in autonomous driving tasks, which requires a sufficient understanding of real-world random interactions and exhibits strong multi-modality. Emerging diffusion models not only demonstrate powerful capabilities in a large number of generation problems but also have excellent performance in pedestrian motion prediction problems. However, the difficulty of learning the interactive information of complex road agents in Markov chains and expensive time consumption prevent the existing diffusion model from in-time prediction and unstable training. To undress this challenge, we propose a novel accelerable goal-conditioned diffusion model.
Motion prediction is a challenging problem in autonomous driving as it demands the system to comprehend stochastic dynamics and the multi-modal nature of real-world agent interactions. Diffusion models have recently risen to prominence, and have proven particularly effective in pedestrian motion prediction tasks. However, the significant time consumption and sensitivity to noise have limited the real-time predictive capability of diffusion models.
%Despite their potential, the complexity of capturing the nuanced interactions among heterogeneous road agents and road conditions, coupled with the constraints of Markov chain processes and sampling time, has impeded the real-time applicability and stability of existing diffusion models. Furthermore, the accuracy of motion predictions is critically undermined by noise interference in sensor data. 
In response to these impediments, we propose a novel diffusion-based, acceleratable framework that adeptly predicts future trajectories of agents with enhanced resistance to noise. The core idea of our model is to learn a coarse-grained prior distribution of trajectory, which can skip a large number of denoise steps. 
%Capitalizing on high-definition map information and the historical trajectories of agents, our method meticulously learns a prior distribution that significantly augments the diffusion process.
This advancement not only boosts sampling efficiency but also maintains the fidelity of prediction accuracy. Our method meets the rigorous real-time operational standards essential for autonomous vehicles, enabling prompt trajectory generation that is vital for secure and efficient navigation. Through extensive experiments, our method speeds up the inference time to 136ms compared to standard diffusion model, and achieves
%1.06\% FDE
significant improvement in multi-agent motion prediction on the Argoverse 1 motion forecasting dataset.

\end{abstract}

%%%%%%%%%%%%%%%%%%%%%%%%%%%%%%%%%%%%%%%%%%%%%%%%%%%%%%%%%%%%%%%%%%%%%%%%%%%%%%%%
\section{INTRODUCTION}

Although autonomous driving systems can significantly enhance convenience, safety remains a major concern. One of the biggest challenges for ensuring safety is accurately predicting the trajectory of vehicles on the road. This involves forecasting future trajectories of modeled agents, taking into account their past movements, other vehicles nearby, road conditions, and traffic signals.

In practice, predicting motion in the real world involves producing multiple future trajectories to capture the probabilistic and multi-modal nature of motion. Some significant progress has been achieved over the past few years\cite{densetnt,gao2020vectornet,zhou2022hivt,gilles2022thomas}. While notable advancements have been made, precisely modeling the stochastic distribution of future trajectories still remains a key challenge due to the multi-modality of trajectories. %Some prior works use the generative models to solve the problem. For instance% 

\begin{figure}[h]
    \centering
    \includegraphics[width=8.5cm]{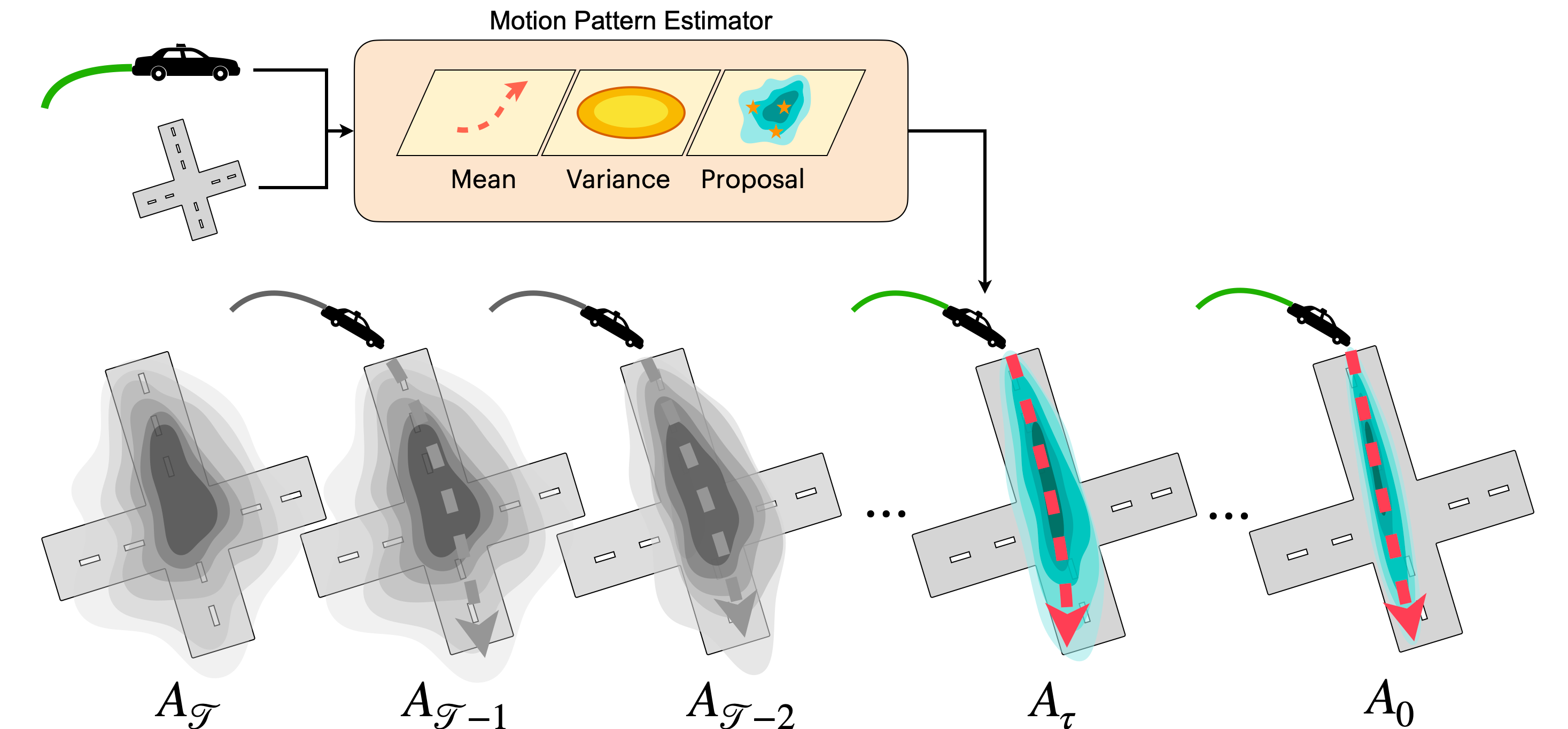} 
    \caption{
        An accelerated diffusion framework featuring a motion pattern estimator for an efficient and effective trajectory denoising process.
    }
    \label{teaser}
    \vspace{-1em}
\end{figure}

Recently, Diffusion models are becoming increasingly popular in the field of machine learning as they serve as powerful tools for various generative tasks\cite{rombach2021highresolution,saharia2022photorealistic,sun2024drivescenegen,feng2023trafficgen,ho2022video,harvey2022flexible}. These models are capable of learning a denoising function based on noisy data and samples from a learned data distribution via iteratively refining a noisy sample starting from pure Gaussian noise and then reverting the process to generate meaningful output from noisy data. However, their high computational costs due to successive denoising and sampling operations limit their performance for real-time applications like motion prediction.
Some prior works use a diffusion model in pedestrian motion prediction tasks and show promising prediction performances[led,mid]. However, vehicle motion prediction tasks in the real world have abundant road constraints and traffic rule information, making the latent space for this task more challenging to learn and model.In this paper, we introduce ADM, an accelerated diffusion-based motion prediction model that is able to meet the real-time inference requirement and concurrently ensures improved robustness. In order to better capture the multi-modality of motion prediction, we leverage the diffusion model. To address prolonged inference times, we present a novel motion pattern estimator, its core idea is to skip a large number of small denoising steps by directly estimating some representative part.
The main contributions of this work are:
\begin{itemize}
    \item We propose a two-stage motion prediction method featuring a diffusion model for denoising agent motions from a learned prior distribution. 
    \item We design a motion pattern estimator that leverages the HD map information and the agent track histories to estimate an explicit prior distribution to facilitate the diffusion process, achieving higher sampling efficiency while not compromising prediction accuracy.
    \item On the Argoverse dataset, our method demonstrates superior performance and robustness against input noise compared to baseline models.
\end{itemize}

\section{Related Works}

\subsection{Denoising Diffusion Probabilistic Models}

Denoising Diffusion Probabilistic Models (DDPM)\cite{DDPM2020,dpm} have indeed emerged as a powerful class of generative models that have been applied successfully across a wide range of domains, such as image generation\cite{rombach2021highresolution, saharia2022photorealistic}, scenario generation\cite{sun2024drivescenegen, feng2023trafficgen}, and audio generation\cite{ho2022video, harvey2022flexible}. 
These models work by gradually transforming noise into structured patterns through a series of learned reverse diffusion steps, typically trained to reverse a Markov chain that gradually adds Gaussian noise to the data. 
However, the application of standard diffusion models in real-time settings is limited due to the requirement of hundreds of denoising steps. To address this, the Denoising Diffusion Implicit Models (DDIM)\cite{song2022DDIM} approach has been developed. DDIM enhances the efficiency of the sampling process by initially reconstructing the original data, and then forecasting the transition to subsequent timestamps via a non-Markovian process, thereby facilitating quicker generation times. Later, a new family of models called Consistency Models\cite{song2023consistency} was proposed, which generates high-quality images by directly mapping noise to data and supports fast one-step generation by design while allowing for multi-step sampling to trade compute for sample quality without explicit training. 

In this work, we develop a motion pattern estimator to estimate a sufficiently expressive distribution. This estimator not only achieves faster inference speed by replacing a large number of former denoising steps but also enables the correlation between samples to adjust sample diversity adaptively, improving diffusion performance in motion prediction.

\subsection{Vehicle Motion Prediction}

Many previous works have attempted to predict agents' trajectories and had impressive achievements in some aspects. THOMAS\cite{gilles2022thomas} presented a unified model architecture for simultaneous agent future heatmap estimation, in which we leverage hierarchical and sparse image generation for fast and memory-efficient inference. HiVT\cite{zhou2022hivt} proposed a translation-invariant scene representation and rotation-invariant spatial learning modules, which extract features robust to the geometric transformations of the scene and model agents by decomposing the problem into local context extraction and global interaction modeling. However, these works didn't sufficiently consider the impact of various modes of perceptual uncertainties, such as noise, missing track history, etc. Motiondiffuser\cite{jiang2023motiondiffuser} proposed a general constrained sampling framework that enables controlled trajectory sampling based on differentiable cost functions. Although they\cite{jiang2023motiondiffuser} indeed reduce sampling time by re-designing the inference method, the traditional diffusion-based motion prediction method has high computational costs due to successive denoising and sampling operations limiting their performance for real-time applications.

\subsection{Diffusion Models for Pedestrian Motion Prediction}

To the best of our knowledge, MID\cite{MID} is the first pedestrian prediction framework using diffusion models to stimulate the indeterminacy of pedestrian motion. Subsequent research has primarily concentrated on enhancing the speed of diffusion-based prediction models and reducing the prediction uncertainty \cite{mao2023leapfrog, autoDM, liu2024intentionaware}.  LeapFrog\cite{mao2023leapfrog} proposed a method to leverage a trainable leapfrog initializer to learn an expressive multi-modal distribution of future trajectories directly. The study in \cite{autoDM} reduces uncertainty by categorizing each pedestrian into their most likely group through a learning process. Furthermore, \cite{liu2024intentionaware} presents an intention-aware denoising diffusion model that separates the original uncertainty into two categories: intention and action， This model employed two interdependent diffusion processes and reduces inference time by diminishing the dimensionality in the intention-aware diffusion stage. Although some impressive results showcase the potential of these methods, due to the differences in scenario modal preference and agent's dynamic model between vehicles and pedestrian motion prediction, a specific designed and novelty model is necessary to fill the gap.

\section{Methodology}

\begin{figure*}[t]
    \centering
    \includegraphics[width=17cm]{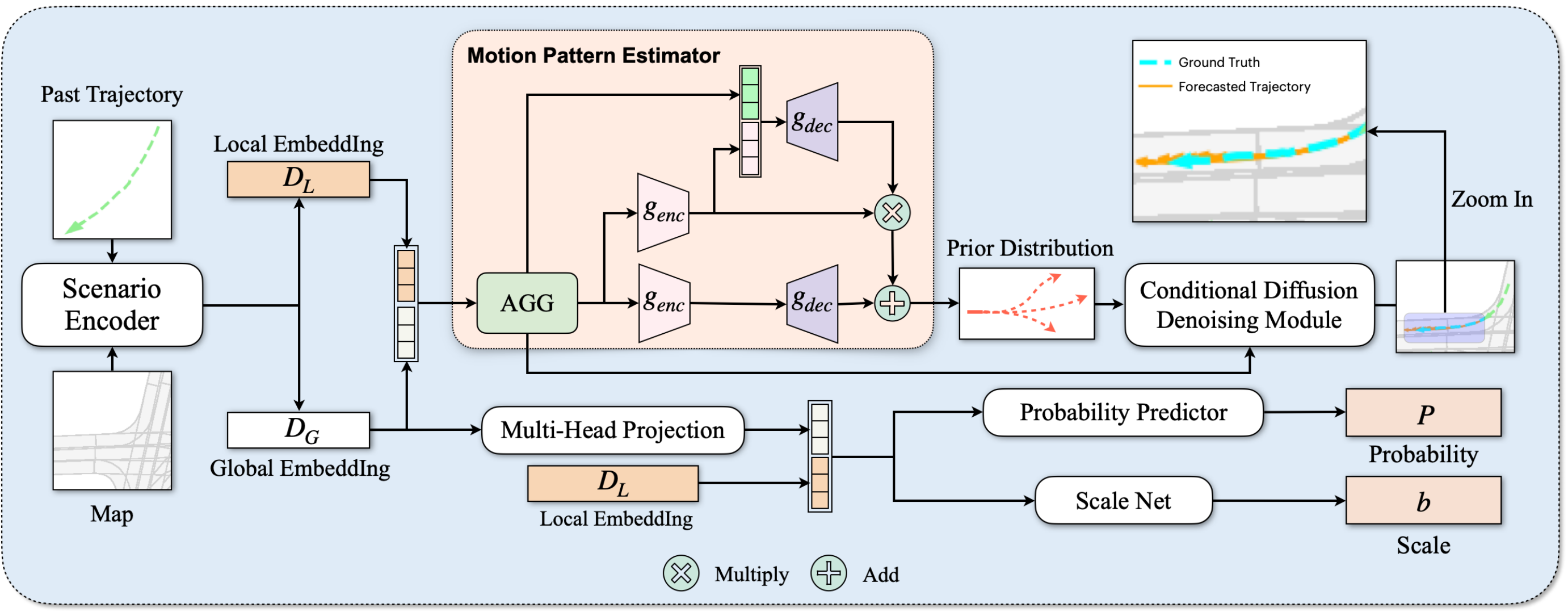} 
    \caption{
      Overview of the pipeline: 1. Scenario Encoder captures interactions among agents and between agents and the map, as well as spatial-temporal relationships in each time frame to gain local embedding and global embedding. 2. Motion Pattern Estimator reparameterizes the prior distribution by learning the mean, variance, and navigation nodes. 3. Conditional Diffusion Denoising Module refines the prior distribution into a clear trajectory. 4. Probability Predictor estimates the probability for each mode of an agent's behavior. 5. Scale Net learns the scale of the Laplace distribution for regression loss.
    }
    \label{fig-overview}
    \vspace{-1em}
\end{figure*}

\subsection{Overview}

% \textcolor{red}{
%     As shown in Fig. \ref{fig-overview}, our proposed method can be divided into three major components, a scenario encoder module, a motion pattern estimator module and a conditional diffusion denoising module.
%     Scenario Encoder captures interactions among agents and between agents and the map, as well as spatial-temporal relationships in each time frame to gain local embedding and global embedding. Motion Pattern Estimator reparameterizes the prior distribution by learning the mean, variance, and navigation nodes. Conditional Diffusion Denoising Module refines the prior distribution into a clear trajectory. Probability Predictor estimates the probability for each mode of an agent’s behavior. Scale Net learns the scale of the Laplace distribution for regression loss.
% }
As shown in Fig. \ref{fig-overview}, our proposed method can be divided into three major components, a Scenario Encoder module, a motion Pattern Estimator Module, and a Conditional Diffusion Denoising Module. Scenario Encoder generates interactions in traffic scenarios into feature embedding. Motion Pattern Estimator learns to model the prior distribution. Conditional Diffusion Denoising refines the prior distribution into a trajectory with determinacy. Additionally, a Probability Predictor calculates the likelihood of each potential action an agent might take. Meanwhile, the Scale Net determines the Laplace distribution’s scale for the regression loss.

First, our method employs an encoder $\mathcal{F}_{encoder}$ that is able to produce two distinct embeddings at the same time: a local embedding $D_{L} \in \mathbb{R}^{N_{a}, D}$ and a global embedding $ D_{G} \in \mathbb{R}^{N_{a}, D}$:
\begin{equation}
    D_{L},D_{G} = \mathcal{F}_{encoder}(\mathbf{A}^{past},\mathbf{M})\\
\end{equation}
where $\mathbf{M}=\{{M}_{1}, {M}_{2}, \ldots, {M}_{N_{m}}\}$ represents the map information, including centerlines, road turn directions, intersections, road boundaries, and traffic controls in the given scenario. 
Meanwhile, $\mathbf{A}^{past} \in \mathbb{R}^{N_{a} \times T_{p} \times 2}$ represents the history trajectory $\mathbf{A}^{past}$ for all the traffic agents in the scenario over the past $T_{p}$ timesteps.
The history trajectory of the $i$-th vehicle can be represented by $\mathbf{A}^{past}_{i}=\{a_{i}^{t-T_{p}}, \ldots, a_{i}^{t}\}$, where $i \in \{1, \dots, N_{a}\}$.

Then, the two embeddings are concatenated together and fed into a linear layer $\varphi_{agg}$ to form an aggregate embedding, which is later fed into the motion pattern estimator module $\mathcal{F}_{estimator}$ to predict mean $ \mu_{\theta} \in \mathbb{R}^{N_{a} \times T_{f} \times 2} $, variance $ \sigma_{\theta} \in \mathbb{R}^{N_{a} \times 1} $ and navigation node $ N_{nodes} \in \mathbb{R}^{N_{a} \times T_{f} \times 2 \times K} $. We then model the top $K$ coarse-grained prior distribution trajectories $ \Tilde{\mathbf{A}} \in \mathbb{R}^{N_{a} \times T_{f} \times 2 \times K} $ by applying parameterization: 
\begin{equation}
    \Tilde{\mathbf{A}}_{{\gamma}} = \mathcal{F}_{estimator}(D_{L} \oplus D_{G}) \\
\end{equation}

Subsequently, these trajectories are refined through finite $\gamma$  iterations by a pretrained conditional denoising diffusion network $\mathcal{F}_{denoising} $, using the embeddings as context, ultimately yielding the future trajectory $\mathbf{A} \in \mathbb{R}^{ N_{a} \times T_{f} \times 2 \times K}$. 
\begin{equation}
    \begin{aligned}
        \Tilde{\mathbf{A}}_{\tau} = \mathcal{F}_{denoising} &(\Tilde{\mathbf{A}}_{\tau+1},\tau, \varphi_{agg} \left( D_{L} \oplus D_{G} \right)) \\
        \tau &\in \{\gamma -1, \ldots, 0\} \\
    \end{aligned}
\end{equation}

Lastly, we use a linear layer $g_{proj}$ on $D_{G}$ and concatenate it with $D_{L}$ as a new embedding. The new embedding then passes through two multilayer perceptrons (MLPs), namely the probability predictor $g_{prob}$ and the Laplace scale net $g_{laplace}$. This process results in the probability of modality $P \in \mathbb{R}^{N_{a} \times K}$ and an auxiliary quantity, Laplace scale $b \in \mathbb{R}^{N_{a} \times T_{f} \times 2}$, which is utilized in the loss function. 
\begin{equation}
    \begin{aligned}
        P &= g_{proj}(D_{L} \oplus g_{proj}(D_{G})) \\
        b &= g_{laplace}(D_{L} \oplus g_{proj}(D_{G})) \\
    \end{aligned}
\end{equation}

\subsection{ Map \& Agent Encoder}
Inspired by \cite{zhou2022hivt}, we partition the space into $N$ local regions, each of which is anchored by a central agent. For the information of local agent $i$, denoted as $D_{L,i} $, We design a cross-attention interaction network to encode the social dynamics between agents as well as between agents and lanes. To capture the temporal dynamics of each agent, we utilize a Gated Recurrent Unit layer. To encode the disparities between different regions, a global attention network conditioned on each agent's local embedding $D_{L}$ is employed to learn the overall interaction patterns among agents. The latent information of each local region and global interaction $D_{L}$ and $D_{G}$ are then aggregated as the embedding token D as conditioned for the Denoiser and motion pattern estimator.
% \begin{equation}
%     \alpha _{t}^{i}=softmax\left( \frac{q_{i}^{t^{T} }}{\sqrt{d_{k}} }\cdot \left [ \left \{ k_{ij}^{k}  \right \}_{j\in N_{i} }   \right ]   \right)  
% \end{equation}

\subsection{Diffusion Process for motion prediction}
The purpose of the forward diffusion process is to erode the information in trajectory $\mathbf{A}_{i,k}$ by gradually adding Gaussian noise and ultimately turning it into a complete Gaussian noise. With each agent's trajectory $\mathbf{A}_{i,k} = \{a_{i,k}^{t+1}, \ldots, a_{i,k}^{t+T_{f}}\}$, we define this forward diffusion process as $\mathbf{A}_{0}, \mathbf{A}_{1}, \cdots, \mathbf{A}_{\mathcal{T}}$, where $\mathcal{T}$ denotes the maximum number of diffusion steps. On the contrary, we use a diffusion model to learn a reverse process $ \mathbf{A}_{\mathcal{T}}, \mathbf{A}_{\mathcal{T}-1}, \cdots, \mathbf{A}_{0} $, gradually removing uncertainty from the complete Gaussian noise to generate a trajectory. The transition process of the forward and reverse diffusion process satisfies the properties of the Markov chain.

For the diffusion process, the posterior distribution of $\mathbf{A}_{0}, \mathbf{A}_{1}, \cdots, \mathbf{A}_{\mathcal{T}}$ are be defined as:
\begin{equation}
    \begin{aligned}
        q\left(\mathbf{A}_{1: \mathcal{T}} \mid \mathbf{A}_{0}\right) &:=\prod_{\tau=1}^{\mathcal{T}} q\left(\mathbf{A}_{\tau} \mid \mathbf{A}_{\tau-1}\right) \\
        q\left(\mathbf{A}_{\tau} \mid \mathbf{A}_{\tau-1}\right) &:=\mathcal{N}\left(\mathbf{A}_{\tau} ; \sqrt{1-\beta_{\tau}} \mathbf{A}_{\tau-1}, \beta_{\tau} \mathbf{I}\right)
    \end{aligned}
\end{equation}
where $\beta_{1},\beta_{2}...\beta_{\tau}$ are fixed variance schedulers that control the scale of the injected noise. 

The result at any noise level $\tau$ in the diffusion process can be calculated as \cite{DDPM2020}:
\begin{equation}
    q\left(\mathbf{A}_{\tau} \mid \mathbf{A}_{0}\right):=\mathcal{N}\left(\mathbf{A}_{\tau} ; \sqrt{\overline{\alpha}_{\tau}} \mathbf{A}_{0},\left(1-\overline{\alpha}_{\tau}\right) \mathbf{I}\right)
\end{equation}
where $\alpha_{\tau}=1-\beta_{\tau}$ and $\overline{\alpha}_{\tau}=\prod_{\tau=1}^{\tau} \alpha_{\tau}$. 

When $\tau$ is large enough, we can approximately consider $\mathbf{A}_{\mathcal{T}} \sim \mathcal{N}(\mathbf{0}, \mathbf{I})$. That indicates the process in which trajectory $\mathbf{A}$ is gradually broken into Gaussian noise. Then, the process that generates a new trajectory from Gaussian noise can be defined as a reverse diffusion process. Given a state feature $\mathbf{D}$ as input, we formulate the reverse diffusion process as:

% , which includes local embedding $\mathbf{D_L}$ and global embedding $\mathbf{D_G}$, learned by a HIVT encoder parameterized by $\vartheta$ with the history trajectories $\mathbf{A}{i}=\{a_{i}^{t-T_{p}},\ldots,a_{i}^{t}\}$ and contextual map information $\mathbf{M}=\{{M}_{1}, {M}_{2}, \ldots, {M}_{total}\} $ 
\begin{equation}
    p_{\theta}\left(\mathbf{A}_{0: \mathcal{T}} \mid \mathbf{D}\right) := p\left(\mathbf{A}_{\mathcal{T}}\right) \prod_{\tau=1}^{\mathcal{T}} p_{\theta}\left(\mathbf{A}_{k-1} \mid \mathbf{A}_{\tau}, \mathbf{D}\right)
\end{equation}
\begin{equation}
    \begin{aligned}
        & p_{\theta}\left(\mathbf{A}_{\tau-1} \mid \mathbf{A}_{\tau}, \mathbf{D}\right) \\ 
        & :=\mathcal{N}\left(\mathbf{A}_{\tau-1} ; \boldsymbol{\mu}_{\theta}\left(\mathbf{A}_{\tau}, \tau, \mathbf{D}\right) ; \boldsymbol{\Sigma}_{\theta}\left(\mathbf{A}_{\tau}, \tau\right)\right)
    \end{aligned}
\end{equation}
% \begin{equation}
%     \begin{aligned}
%         p_{\theta}\left(\mathbf{A}_{0: \mathcal{T}} \mid \mathbf{D}\right) & :=p\left(\mathbf{A}_{\mathcal{T}}\right) \prod_{\tau=1}^{\mathcal{T}} p_{\theta}\left(\mathbf{A}_{k-1} \mid \mathbf{A}_{\tau}, \mathbf{D}\right) \\ p_{\theta}\left(\mathbf{A}_{\tau-1} \mid \mathbf{A}_{\tau}, \mathbf{D}\right) & := \mathcal{N}\left(\mathbf{A}_{\tau-1}; \boldsymbol{\mu}_{\theta}\left(\mathbf{A}_{\tau}, \tau, \mathbf{D}\right); \boldsymbol{\Sigma}_{\theta}\left(\mathbf{A}_{\tau}, \tau\right) \right)
%     \end{aligned}
% \end{equation}
where $p\left(\mathbf{A}_{\mathcal{T}}\right)\sim \mathcal{N}(\mathbf{0}, \mathbf{I})$ denotes the initial noise that sample from Gaussian distribution. $\theta$ denotes the parameters of diffusion model, which are acquired by training with existing agents' trajectories. 

The variance term of the Gaussian transition can be set as $\boldsymbol{\Sigma}_{\theta}\left(\mathbf{A}_{\mathcal{T}}, k\right)=\sigma_{\tau}^{2} \mathbf{I}=\beta_{\tau} \mathbf{I}$. As shown in \cite{DDPM2020}, $\boldsymbol{\mu}_{\theta}\left(\mathbf{A}_{\tau}, \tau\right)$ is obtained by calculating $\mathbf{A}_{\mathcal{\tau}}$ and model output, which follows the formula below：

\begin{equation}
    \boldsymbol{\mu}_{\theta}\left(\mathbf{A}_{\tau}, \tau\right)=\frac{1}{\sqrt{\alpha_{k}}}\left(\mathbf{A}_{\tau}-\frac{\beta_{\tau}}{\sqrt{1-\overline{\alpha}_{\tau}}} \epsilon_{\theta}\left(\mathbf{A}_{\tau}, \tau\right)\right)
\end{equation} 
Although the standard diffusion model is expressively powerful in learning sophisticated distributions, the running time of a diffusion model is constrained by the large number of denoising steps, which make it fail to achieve real-time inference. Meanwhile, less denoising steps usually cause a weaker representation ability of future distributions. To achieve higher efficiency while preserving a promising representation ability, we propose our specific diffusion model, that includes a trainable motion pattern estimator to capture sophisticated distributions which substitutes a large number of denoising steps.

\subsection{ Motion Pattern Estimator.}
%In this section, we provide a comprehensive detail of the specific structure of Motion Pattern Estimator. 

Motion Pattern Estimator skips a large number of small denoising steps by directly estimating some representative parts, which can significantly accelerate the inference speed without losing representation ability. Specifically, considering the total denoised steps using DDPM is $\mathcal{T}$, our goal is to identify a latent space mapping relationship at denoised time $\tau = \gamma$, within the sophisticated latent space constructed by the forward diffusion process, which closely approximates the true distribution at denoised steps $\tau=0$. Our experiments indicate that by directly predicting the prior data distribution at denoised steps $\tau = \gamma$, we can significantly enhance inference speed while maintaining performance improvements. 

Our motion pattern estimator consists of three modules, two of then estimate the mean  $\mu $ and variance $\sigma $, and the last one estimates some navigation nodes. Assume $D_{L} $ and $D_{G} $ to be the local and global embedding tokens of past trajectory and map information. The mean is obtained as follows,
\begin{equation}
    \mu _{\theta } = g_{dec}\left( g_{enc}\left(  \varphi _{agg} \left( D_{L} \oplus D_{G} \right) \right)   \right) 
\end{equation}

Where the function $\varphi _{agg} \left( \cdot  \right)$ aggregates local information and global dynamic connection, $g_{enc}\left( \cdot  \right) $ transforms the individual agent feature information to latent space, and $g_{dec}\left( \cdot  \right) $ creates a mapping from latent space to estimated mean. In practice, $g_{enc}\left( \cdot  \right) $ is a multi-layer perceptron(MLP), specifically, the MLP contains a single fully connected layer followed by layer normalization and then ReLU non-linearity. The $g_{dec}\left( \cdot  \right) $ is also a multi-layer perceptron, while the $\varphi _{agg} \left( \cdot  \right)$ uses a simple concatenation followed by a multi-layer perceptron. The other two modules also share the same structure as the mean prediction module. Note that our navigation nodes prediction module is slightly different from others, it takes the estimated variance as input, formulated as
\begin{equation}
    \sigma_{\theta } = g_{enc} \left(  \varphi _{agg}\left( D_{L} \oplus D_{G} \right) \right)
\end{equation}
\begin{equation}
    N_{nodes} =g_{dec} \left( \varphi_{agg} \left( D_{L} \oplus D_{G}   \right),\sigma_{\theta } \right) 
\end{equation}
where $\sigma_{\theta }$ is the estimated variance, and $N^{nodes} $ is the estimated navigation nodes.
After obtaining $\gamma$ samples 
%$ \mathbf{A}_{\mathcal{T}}, \mathbf{A}_{\mathcal{T}-1}, \cdots, \mathbf{A}_{\tau} $
from motion pattern estimator, we execute the remaining $\mathcal{T} - \tau$ denoising steps to iteratively refine those predicted trajectories.

\subsection{ Conditional Diffusion Denoising Module}

We propose a specific diffusion module to refine trajectories that are predicted by a motion pattern estimator. Different from the standard diffusion model, our module denoises trajectories from a prior distribution $\Tilde{\mathbf{A}}_{\tau,k}$, which calculates from estimated motion pattern: $\mu_{\theta}$, $\sigma_{\theta }$ and $N_{nodes} $. 

\begin{equation}
    \Tilde{\mathbf{A}}_{\tau,k} =\mu _{\theta } +\sigma _{\theta } \cdot N_{nodes} 
\end{equation}

Inspired by \cite{MID}, we define a transformer-based diffusion architecture to model the Gaussian transitions in Markov chain denoting as $\mathcal{F}_{denoising}$, taking noisy trajectories, noise level $\tau$ and embedding $\varphi_{agg} \left( D_{L} \oplus D_{G} \right)$ as input. The overall procedure of the proposed diffusion model is formulated in Algorithm \ref{algo-inference}.

\begin{algorithm}[h]
\caption{Model Inference}
\label{algo-inference}
\begin{algorithmic}[1]
    \algsetup{linenosize=\scriptsize}
    \small
    \REQUIRE ~~\\
        Past trajectories $\mathbf{A}^{past}$; map context $\mathbf{M}$; accelerated denoising steps $\gamma$. \\
    \ENSURE ~~\\
        Predicted trajectories $\hat{\mathbf{A}}$; probabilities $\mathbf{P}$. \\
        \STATE $ D_{L},D_{G} \gets \mathcal{F}_{encoder}(\mathbf{A}^{past},\mathbf{M})$   
        \STATE $\mathbf{\mu} _{\theta } \gets g_{dec}\left( g_{enc}\left( \varphi _{agg} \left( D_{L} \oplus D_{G} \right)  \right) \right)$ \quad \COMMENT{mean}
        \STATE $\sigma_{\theta } \gets g_{enc} \left(\varphi _{agg}\left( D_{L} \oplus D_{G} \right) \right)$ \quad \COMMENT{variance}
        \STATE $N_{nodes}  \gets g_{dec} \left( \varphi _{agg}\left( D_{L} \oplus D_{G} \right),\sigma_{\theta} \right) $ \quad \COMMENT{estimated navigation nodes}
        \STATE $\mathbf{P} \gets g_{prob}(D_{L} \oplus g_{proj}(D_{G}))$
        \STATE $\Tilde{\mathbf{A}}_{\tau,k} \gets \mu_{\theta} + \sigma_{\theta} \cdot N_{\text{nodes},k}$ , $k \in \{1, \ldots, K\}$
        \FORALL{ $\tau \in \{\gamma - 1, \ldots, 0\}$ }
            \STATE $\Tilde{\mathbf{A}}_{\tau,k} \gets \mathcal{F}_{denoising}(\Tilde{\mathbf{A}}_{\tau+1,k},\tau, \varphi_{agg} \left( D_{L} \oplus D_{G} \right))$
        \ENDFOR
        \STATE $\hat{\mathbf{A}} \gets \{\Tilde{\mathbf{A}}_{0,1}, \ldots,\Tilde{\mathbf{A}}_{0,K}\}$ 
        \RETURN $\hat{\mathbf{A}},\mathbf{P} $
\end{algorithmic}
\end{algorithm}

\begin{figure*}[t]
    \centering
    \includegraphics[width=0.99\linewidth]{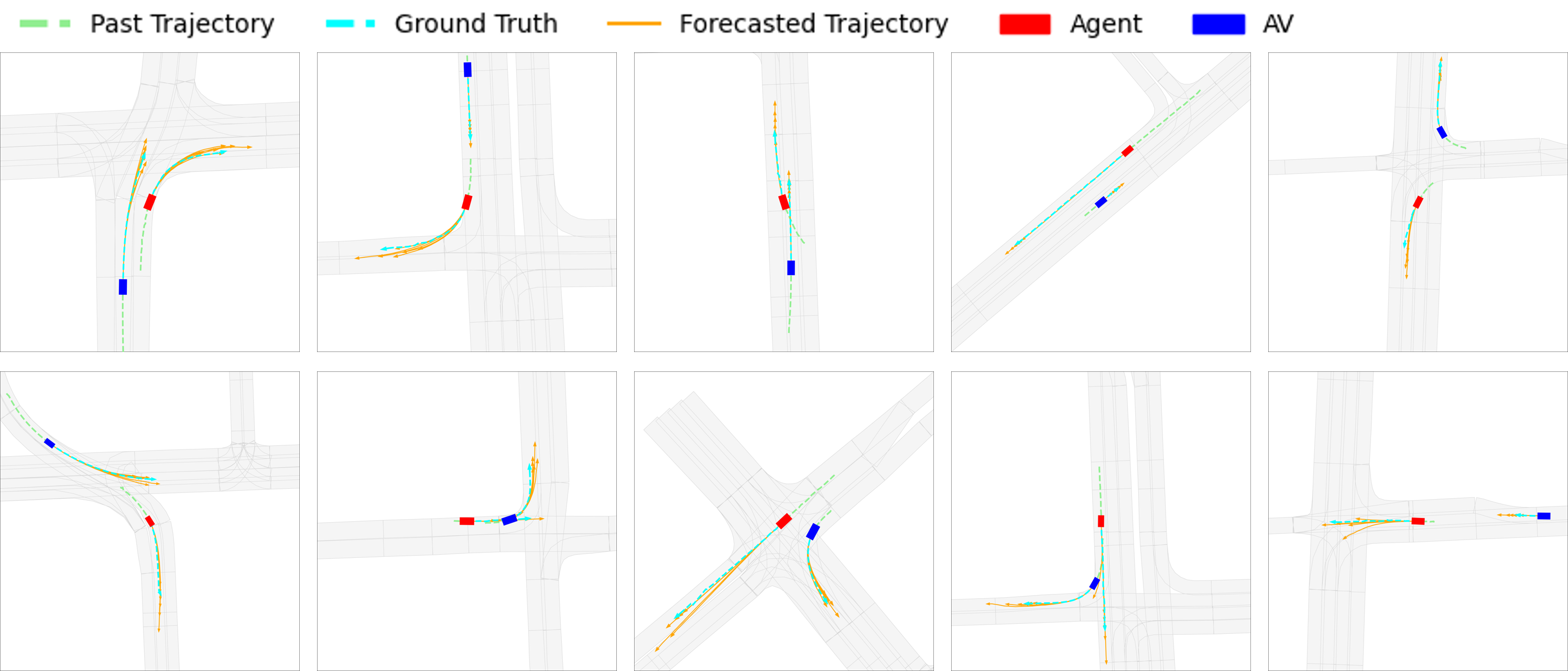} 
    \caption{
      Qualitative evaluation of the ADM. The prediction samples show high accuracy in multi-agent motion prediction on the Argoverse 1 motion forecasting dataset.
    }
    \label{fig-qualitative}
\end{figure*}

Compared to the standard diffusion model, our motion pattern estimator directly models the $\tau$th denoised distribution $p_{\theta}(\mathbf{A}_{\tau})$, which is hypothetically equivalent to the output of executing ($\mathbf{T} - \tau$) denoising steps. We then draw samples from the distribution $p_{\theta}(\mathbf{A}_{\tau})$ and obtain $K$ future trajectories $\mathbf{A}_{a,k} = \{a_{a,k}^{t+1},\ldots,a_{a,k}^{t+T_{f}}\}$, where $k$ means $k$ samples are dependent to intentionally allocate appropriate sample diversity.

\subsection{Loss Function}
The model undergoes a two-phase training. Initially, the conditional diffusion module is directly attached to the backend of the scenario encoder, serving as a decoder. This configuration is employed to train the diffusion model’s denoising ability, taking into account the context provided by local and global embeddings. After calculating the $\mathcal{L}_{2}$ error(the Euclidean distance between predictions and true targets) with ground truth trajectories, we mark the predicted trajectory with minimum $\mathcal{L}_{2}$ error as $\hat{\mathbf{A}}_{best}$. Negative log-likelihood($NLL$) loss is the metric of choice during this phase to optimize the performance of noisy trajectory refinement.
\begin{equation}
\begin{split}
\text{NLL}(\mathbf{A}, \hat{\mathbf{A}}_{best}, b) = 
\frac{|\mathbf{A} - \hat{\mathbf{A}}_{best} | }{b} + \log(2b) 
\end{split}
\end{equation}
where $\hat{\mathbf{A}}_{best}$ is the best predicted trajectory and $b$ is the scale learned by Laplace scale net $ g_{laplace} $. 

In the subsequent phase, the parameters of the conditional diffusion module are frozen, shifting the focus to enhancing the motion pattern estimator’s proficiency in generating coarse-grained trajectories. 
At this phase, the loss function of the model encompasses $NLL$ loss for trajectory regression and cross-entropy loss for modality classification. It first calculates $\mathcal{L}_{2}$ errors of predicted trajectories and generates them into soft target probabilities distribution denoting as $st \in \mathbb{R}^{a\times K} $. This transforms smaller $\mathcal{L}_{2}$ error values (trajectory closer to the target) into larger probabilities, and larger $\mathcal{L}_{2}$ error values into smaller probabilities. The aim is to optimize the ability of probability predictor $ g_{prob}$ to generate higher scores for more accurate predicted trajectories.  
\begin{equation}
    \text{$\mathcal{L}_{CE}$}(st, \mathbf{P}) = -\frac{\sum_{i=1}^{N_{a}} st_{i} \log(softmax(\mathbf{P}_{i}))}{N_{a}}
\end{equation}
where $st$ represents the soft target probabilities, $\mathbf{P}$ represents the estimated scores.

\section{EXPERIMENTS}

\begin{table*}[htbp]
\centering
\caption{\href{https://eval.ai/web/challenges/challenge-page/454/leaderboard/}{Leadboard}  on Argoverse Motion Forecasting Dataset.}
\label{tab:marginal}
\begin{tabular}{cccccc} 
    \hline
    \rowcolor[rgb]{0.863,0.863,0.863}  Method ~& Reference & minADE $\downarrow$ & minFDE $\downarrow$ & Miss Rate $\downarrow$ & brief-minFDE $\downarrow$ \\
    \hline
    \\[-1em]
    GOHOME\cite{gilles2021gohome} & ICRA 2021 & 0.9425 & 1.4503 & 0.1048 & 1.9834 \\
    THOMAS\cite{gilles2022thomas} & ICLR 2022 & 0.9423 & 1.4388 & \textbf{0.1038} & 1.9736 \\
    Dense TNT\cite{densetnt} & ICCV 2021 & 0.8817 & 1.2815 & 0.1258 & 1.9759 \\
    mmTransformer\cite{liu2021multimodal} & CVPR 2021 & 0.8436 & 1.3383 & 0.1540 & 2.0328  \\
    ssl-lane\cite{bhattacharyya2022ssllanes} & CoRL 2022 & 0.8401 & 1.9433 & 0.1326 & 1.9433  \\
    FRM\cite{park2023leveraging} & ICLR 2023 & 0.8165 & 1.2671 & 0.1430 & 1.9365  \\
    TPCN\cite{ye2021tpcn} & CVPR 2021 & 0.8153 & 1.2442 & 0.1333 & 1.9286  \\
    Holistic transformer\cite{hu2022holistic} & PR 2023 & 0.8123 & 1.2227 & 0.1303 & 1.9172  \\
    HiVT-128-checkpoints\cite{zhou2022hivt} & CVPR 2022 & 0.7994 & 1.2321 & 0.1369 & \textbf{1.9010}  \\
    \rowcolor[rgb]{0.863,0.863,0.863} Ours & \_ &\textbf{0.7916} & \textbf{1.2191} & 0.1409 & 1.9136  \\
    \hline
\end{tabular}
\end{table*}

\subsection{Experimental Setup}
\subsubsection{Dataset} To assess the effectiveness of our proposed approach, we employed the {\bf Argoverse 1 motion forecasting dataset} \cite{chang2019argoverse}, renowned for its provision of comprehensive agent trajectories and high-resolution map data. This dataset encompasses 205,942, 39,472, and 78,143 sequences allocated for training, validation, and testing, respectively. Notably, each sequence is uniformly sampled at a rate of 10 Hz, with the core task revolving around the prediction of future 3-second trajectories based on 2 seconds of historical observational data. In the test set, only the first 2-second trajectories are available. 
\subsubsection{Metrics}  Our evaluation methodology employs standard metrics for motion prediction, including minimum Average Displacement Error $(minADE_{k})$, minimum Final Displacement Error $(minFDE_{k})$, Miss Rate $(MR_{k})$, and $brier-minFDE_{k}$. These metrics systematically assess the optimal predicted trajectory among K hypothetical trajectories for a single target agent, comparing them against the ground truth. The $minADE_{k}$ quantifies the Euclidean distance between the optimal predicted trajectory and the ground truth trajectory, averaged over prospective time intervals. Conversely, $minFDE_{k}$ evaluates the discrepancy solely at the terminal time step, while $MR_{k}$ delineates the ratio of scenarios where the deviation between the terminal points of the ground truth trajectory and the optimal predicted trajectory exceeds a threshold of 2.0 meters. The $brier-minFDE_{k}$ integrates a brier score $\left( 1-P \right) ^{2}$ with $minFDE_{k}$, where P represents the probability associated with the optimal predicted trajectory.
\subsubsection{Implementation details}  We train our model for 64 epochs on three RTX 3090 GPUs using AdamW optimizer\cite{loshchilov2019decoupled}. The scenario encoder consists of 1 layer of agent-agent and agent-lane interaction module, 4 layers of temporal dynamic learning module, and 3 layers of global attention module. The number of heads in all multi-head attention blocks is 8. The radius of all local regions is 50 meters. The hidden size of our embedding is 128. To train our diffusion model, we set the diffusion steps $\mathcal{T}$ = 1000, and the estimator step $\gamma$ = 5. In the denoising module, we build our core transformer-based denoising network with a hidden size of 128. With a frozen denoising module, we then train our motion pattern estimator for 64 epochs with an initial learning rate of $5\times 10^{-4}$.

\subsection{Experiment Results}

%\begin{table}[htbp]
%\caption{Inference time and performance in different sample %method.}
%\label{tab:Inference time}%
%\centering
%\begin{tabular}{cccc}
    %\hline
    %\rowcolor[rgb]{0.863,0.863,0.863} Sampling method~& Step & K %&Inference(ms) $\downarrow$ \\ 
    %\hline
    %\\[-1em]
    %DDPM & 1000 & 6 & 11578.92 \\
    %\hline
    %\\[-1em]
    %\multirow{4}{*}{DDIM} & 50 & 6 & 671.75 \\
    % & 30 & 6 & 453.13 \\
    % & 20 & 6 & 322.80 \\
    % & 5 & 6 & \textbf{136.04} \\
    %\hline
    %\\[-1em]
    % HiVT-128 & none & 6 & 0.6611 & 0.9692 & 0.09203 & 51.33 \\
    % \hline
   %{\cellcolor[rgb]{0.902,0.902,0.902}}Ours & {\cellcolor[rgb]%{0.902,0.902,0.902}} 5 &{\cellcolor[rgb]{0.902,0.902,0.902}} 6 & %{\cellcolor[rgb]{0.902,0.902,0.902}} 136.74 \\
%    \hline
%\end{tabular}
%\end{table}

\begin{table}[htbp]
\caption{Comparison of different sampling methods' prediction performance and inference time. ($K = 6$) }
\label{tab:Inference time}%
\centering
\begin{tabular}{ccccccc}
    \hline
    \rowcolor[rgb]{0.863,0.863,0.863} Method~& Steps & minADE $\downarrow$ & minFDE $\downarrow$ & Miss Rate $\downarrow$ & \begin{tabular}[c]{@{}c@{}}Time $\downarrow$ \\ (ms)\end{tabular} \\ 
    \hline
    \\[-1em]
    DDPM & 1000 & 1.170 & 2.189 & 0.332 & 11578.9 \\
    \hline
    \\[-1em]
    \multirow{4}{*}{DDIM} & 50 & 1.114 & 1.987 & 0.306 & 671.8 \\
     & 30 & 1.156 & 2.089 & 0.326 & 453.1 \\
     & 20 & 1.171 & 2.141 & 0.337 & 322.8 \\
    & 5 & 4.395 & 10.753 & 0.730 & \textbf{136.0} \\
    \hline
    \\[-1em]
    % HiVT-128 & none & 6 & 0.6611 & 0.9692 & 0.09203 & 51.33 \\
    % \hline
    {\cellcolor[rgb]{0.902,0.902,0.902}}Ours & {\cellcolor[rgb]{0.902,0.902,0.902}} 5 & {\cellcolor[rgb]{0.902,0.902,0.902}} \textbf{0.616} &
    {\cellcolor[rgb]{0.902,0.902,0.902}} \textbf{0.875} & {\cellcolor[rgb]{0.902,0.902,0.902}} \textbf{0.082} & {\cellcolor[rgb]{0.902,0.902,0.902}} 136.7 \\
    \hline
\end{tabular}
\end{table}

% \begin{table}[htbp]
% \caption{Inference time and performance of different sampling methods. \\Sample trajectories K=6}
% \label{tab:Inference time}%
% \centering
% \begin{tabular}{ccccc}
%     \hline
%     \rowcolor[rgb]{0.863,0.863,0.863} Method~& Step & minADE $\downarrow$ & minFDE $\downarrow$ &inference(ms) $\downarrow$ \\ 
%     \hline
%     \\[-1em]
%     DDPM & 1000 & 1.1703 & 2.1892 & 11578.92 \\
%     \hline
%     \\[-1em]
%     \multirow{4}{*}{DDIM} & 50 & 1.1136 & 1.9871 & 671.75 \\
%      & 30 & 1.1559 & 2.0892 & 453.13 \\
%      & 20 & 1.1712 & 2.1408 & 322.80 \\
%     & 5 & 4.3951 & 10.753 & 136.04 \\
%     \hline
%     \\[-1em]
%     % HiVT-128 & none & 6 & 0.6611 & 0.9692 & 0.09203 & 51.33 \\
%     % \hline
%     {\cellcolor[rgb]{0.902,0.902,0.902}}Ours & {\cellcolor[rgb]{0.902,0.902,0.902}} 5 & {\cellcolor[rgb]{0.902,0.902,0.902}} 0.6156 &
%     {\cellcolor[rgb]{0.902,0.902,0.902}} 0.8746 & {\cellcolor[rgb]{0.902,0.902,0.902}} 136.74 \\
%     \hline
% \end{tabular}
% \end{table}

The experiment results of our method on the Argoverse benchmark are summarised in TABLE \ref{tab:marginal}. 
It can be observed that, compared to other methods on the Argoverse leaderboard, our model performs better in terms of minADE and minFDE metrics. 
Meanwhile, another experiment on model inference time is also conducted across different sampling methods to validate the efficiency of our sampling approach. For this experiment, a baseline model is created by attaching our diffusion denoising module to the end of a scenario encoder during the training stage \uppercase\expandafter{\romannumeral1} to train the standard denoising process. TABLE \ref{tab:Inference time} presents a comparison between the baseline model with varying numbers of sampling steps using DDIM and DDPM sampling methods and our sampling method in terms of performance and the time taken to infer $K$=6 trajectories for an agent on Argoverse valid set. The results of the comparison demonstrate that our method, by modeling the distribution at the re-parameterized fifth step from the last in the denoising process, achieves enhanced performance while maintaining an inference time roughly equivalent to that of the DDIM method with five inference steps.

\subsection{Robustness Analysis}
In our pursuit to evaluate the durability of our model in noisy conditions, we embarked on robustness experiments. This involved the introduction of Gaussian noise centered at zero with standard deviations spanning from 0 to 1 in increments of 0.2, into our validation dataset. The objective was to measure our model’s ability to sustain its predictive accuracy amidst various intensities of noise. Such an assessment is vital for applications in the real world, where data is often less than perfect. As shown in Fig. 4, The results of the robustness study showcase the insensitivity and steadfastness of our method under different noise disturbances. 
\begin{figure}[h]
    \centering
    \includegraphics[width=0.99\linewidth]{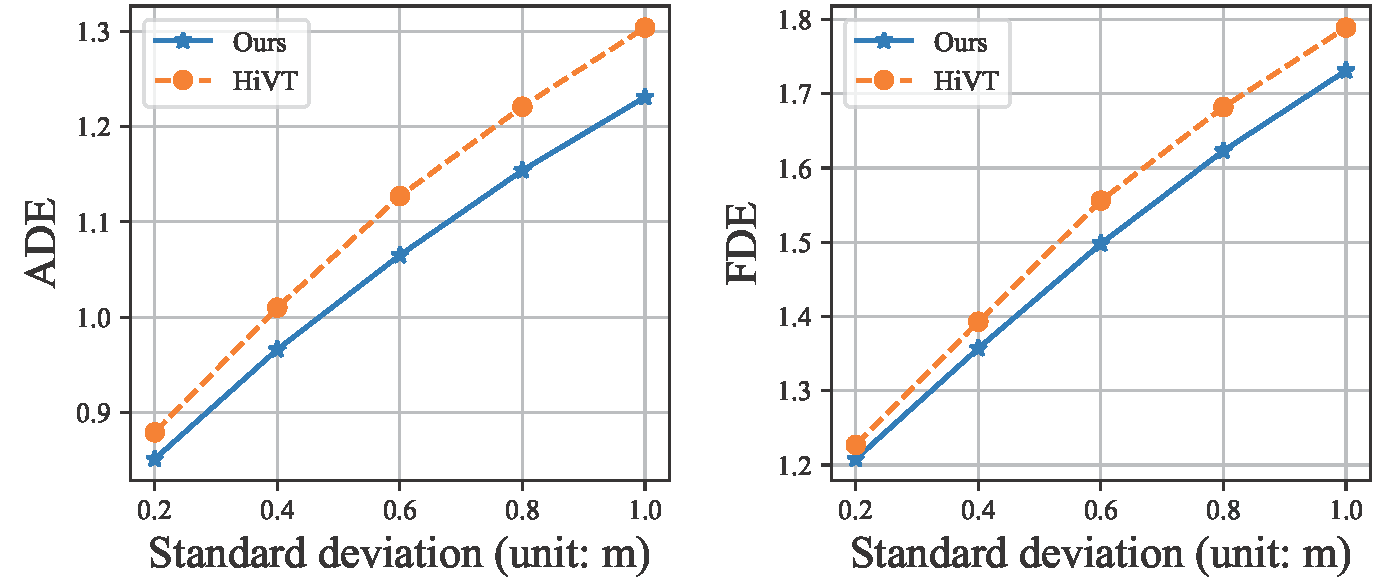} 
    \caption{
       The effect on the model's prediction performance under different noise levels (standard deviations). 
    }
    \label{robust}
    \vspace{-1em}
\end{figure}

\begin{table}[htbp]
\caption{The effect of the motion pattern estimator on the 20\% training set.}
\label{tab:Ablation Study}%
\centering
\begin{tabular}{cccccc}
    \hline
    \rowcolor[rgb]{0.863,0.863,0.863} Estimator ~& Params & K & minADE $\downarrow$ & minFDE $\downarrow$ & Miss Rate $\downarrow$ \\ 
    \hline
    \\[-1em]
    \multirow{3}{*}{\XSolidBrush} & \multirow{3}{*}{\_} & 6 & 1.1559 & 2.0892 & 0.3259 \\
    && 20 & 0.9131 & 1.3151 & 0.1746 \\
    && 40 & 0.8293 & 1.0294 & 0.1261 \\
    \hline
    \\[-1em]
    MLP & 230k & 6 & 0.7879 & 1.3613 & 0.1893\\
    \hline
    \\[-1em]
     \checkmark & 1.6M & 6 & \textbf{0.6681} & \textbf{0.9714} & \textbf{0.0952} \\
    \hline
    \\[-1em]
\end{tabular}
\end{table}
%(Without motion pattern predictor,sample from i.i.d Gaussian noise using DDIM)

\subsection{Ablation Study}

To ensure the effective predictive capability of our motion pattern estimator, we conducted an ablation study on the motion pattern estimator module using 20\% of the training dataset. In this study, $K$ represents the number of predicted trajectories. We configured our baseline model with the employment of the DDIM sampling method from an independent and identically distributed ($i.i.d$) source. Subsequently, we integrated a HiVT decoder between the scenario encoder and the conditional diffusion denoising module to evaluate the performance of a multilayer perceptron (MLP) with 230k parameters. Finally, in Training Stage \uppercase\expandafter{\romannumeral2}, we replaced the HiVT decoder with our unique motion pattern estimator, which has 1.6M parameters. The results of the study are recorded in TABLE \ref{tab:Ablation Study}.

\section{CONCLUSIONS}

% summarize work done, summarize contributions abd results, mention future work 
This paper presents ADM, an accelerated diffusion-based motion prediction model, which achieves better robustness against noise disturbance and significantly speeds up the inference time by specially designing a motion pattern estimator to learn a prior distribution of trajectory. Extensive experimental results show that our model performs better than the other baseline methods on the Argoverse benchmark. Meanwhile, our method is able to achieve faster inference speed, meet the real-time requirement, and demonstrate more robust performance under perceptual uncertainties. 

% In this work, we learn to model the coarse-grained prior distribution of trajectories, however, when changing to other more sophisticated datasets, the feasibility still needs to be proved, and to improve the generalization ability of our model, we can use the classifier-guidance method to help us better model the multimodality of the trajectory.

\addtolength{\textheight}{-12cm}   % This command serves to balance the column lengths

%%%%%%%%%%%%%%%%%%%%%%%%%%%%%%%%%%%%%%%%%%%%%%%%%%%%%%%%%%%%%%%%%%%%%%%%%%%%%%%%

%%%%%%%%%%%%%%%%%%%%%%%%%%%%%%%%%%%%%%%%%%%%%%%%%%%%%%%%%%%%%%%%%%%%%%%%%%%%%%%%



% \section*{ACKNOWLEDGMENT}

%%%%%%%%%%%%%%%%%%%%%%%%%%%%%%%%%%%%%%%%%%%%%%%%%%%%%%%%%%%%%%%%%%%%%%%%%%%%%%%%

\bibliographystyle{IEEEtran}
\bibliography{root}

\end{document}